\documentclass[11pt]{article}
\usepackage[letterpaper,margin=1in]{geometry}

\usepackage{ifpdf}
\usepackage[section]{placeins}
\usepackage[T1]{fontenc}
\usepackage[utf8]{inputenc}
\usepackage{amsfonts}
\usepackage{graphicx}
\usepackage{epstopdf}
\usepackage{enumitem}
\usepackage{algorithmic}
\usepackage{booktabs}
\usepackage{float}
\usepackage{amssymb}
\usepackage{mathtools}
\usepackage{amsthm}
\usepackage{caption}
\usepackage{xcolor}
\usepackage{tikz}
\usepackage{authblk}
\usepackage[hidelinks]{hyperref}
\usepackage[capitalize,noabbrev]{cleveref}
\usetikzlibrary{arrows.meta, shapes.geometric, positioning, calc}
\ifpdf
  \DeclareGraphicsExtensions{.eps,.pdf,.png,.jpg}
\else
  \DeclareGraphicsExtensions{.eps}
\fi

\theoremstyle{remark}

\crefname{hypothesis}{Hypothesis}{Hypotheses}
\theoremstyle{plain}

\usepackage{amsopn}

\begin{document}

\newcommand\relatedversion{}
\renewcommand\relatedversion{\thanks{The full version of the paper can be accessed at \protect\url{https://arxiv.org/abs/0000.00000}}}

\title{TSN-Affinity: Similarity-Driven Parameter Reuse for Continual Offline Reinforcement Learning}

\author[1]{Dominik Żurek}
\author[1]{Kamil Faber}
\author[1]{Marcin Pietron}
\author[1]{Paweł Gajewski}
\author[2]{Roberto Corizzo}
\affil[1]{AGH University of Krakow, Poland}
\affil[2]{American University, Washington, DC, USA}
  
\date{}

\maketitle
\thispagestyle{empty}



\begin{abstract}
Continual offline reinforcement learning (CORL) aims to learn a sequence of tasks from datasets collected over time while preserving performance on previously learned tasks. This setting corresponds to domains where new tasks arise over time, but adapting the model in live environment interactions is expensive, risky, or impossible. However, CORL inherits the dual difficulty of offline reinforcement learning and adapting while preventing catastrophic forgetting. Replay-based continual learning approaches remain a strong baseline but incur memory overhead and suffer from a distribution mismatch between replayed samples and newly learned policies. At the same time, architectural continual learning methods have shown strong potential in supervised learning but remain underexplored in CORL.
In this work, we propose TSN-Affinity, a novel CORL method based on TinySubNetworks 
and Decision Transformer.
The method enables task-specific parameterization and controlled knowledge sharing through a RL-aware reuse strategy that routes tasks according to action compatibility and latent similarity.
%
We evaluate the approach on benchmarks based on Atari games and simulations of manipulation tasks with the Franka Emika Panda robotic arm, covering both discrete and continuous control. Results show strong retention from sparse SubNetworks, with routing further improving multi-task performance. 
Our findings suggest that similarity-guided architectural reuse is a strong and viable alternative to replay-based strategies in a CORL setting. 
Our code is available at: \url{https://github.com/anonymized-for-submission123/tsn-affinity}.
\end{abstract}

\section{Introduction}
\label{sec:introduction}

{\color{black}
Continual learning and reinforcement learning address two complementary requirements of adaptive systems: the ability to accumulate knowledge over time and the ability to make optimized sequential decisions \cite{abel2023definition}. Reinforcement learning provides a principled framework for learning such decision policies by optimizing cumulative reward through interaction with an environment. However, in many real-world applications, the interaction with the environment is expensive, unsafe, or infeasible, which requires learning from previously collected data \cite{offline}. This setting has motivated significant progress in offline reinforcement learning (offline RL), where policies are learned entirely from gathered datasets without further environment interaction.

Despite advances in offline RL, the CORL setting, in which the model adapts to arriving tasks, remains relatively underexplored \cite{gai2023oer}. In CORL, the learner must incorporate new knowledge while preserving performance on previously learned tasks, without access to past data unless explicitly stored \cite{gai2023oer}. This setting combines two major challenges: i) tackling out-of-distribution exploration and distributional shift between training and evaluation, typical of offline RL, and ii) preventing catastrophic forgetting of previous tasks while learning new ones, typical of continual learning.

Recent developments in offline RL show that sequence modeling approaches such as the Decision Transformer or Trajectory Transformer \cite{chen2021decisiontransformer, janner2021trajectorytransformer} offer a flexible and stable framework for learning from static trajectories. Because this paradigm reformulates RL as a supervised sequence problem, it serves as an excellent foundation for isolating and studying continual learning dynamics.

Existing approaches to continual RL, including those adapted to the offline setting, are predominantly replay-based, relying on storing and reusing past samples to mitigate forgetting  \cite{gai2023oer,zhang2023recall,rolnick2019experience}. In contrast, in domains such as continual image classification and anomaly detection, architectural approaches have demonstrated strong resistance to forgetting \cite{wang2024comprehensive,202601.1931}.
Methods such as PackNet \cite{mallya2018packnet}, Piggyback \cite{mallya2018piggyback}, and HAT \cite{serra2018hat} demonstrate that task-specific masks or gating mechanisms can preserve prior knowledge by limiting destructive interference across tasks. 
TinySubNets \cite{pietron2025tsn} extends these ideas with weight sharing, replay memory, and KL-based task routing.
However, the applicability of such methods to CORL is non-trivial, as the transition from supervised learning to reinforcement learning introduces additional challenges, including sequential decision dependencies, policy-induced distributional shift, and the need to generalize across temporally extended trajectories. In particular, effective reuse of knowledge across task-specific subnetworks is critical, yet fundamentally more challenging in reinforcement learning. 


 Building on this, we propose \textbf{TSN-Affinity}, a continual learning method for CORL that introduces a similarity-driven parameter reuse paradigm for sequence-based reinforcement learning. Built on sparse task-specific TinySubNetworks \cite{pietron2025tsn}, TSN-Affinity leverages the Decision Transformer \cite{chen2021decisiontransformer} architecture.
Central to our approach is \textbf{Affinity Routing}, a CORL-specific mechanism that dynamically selects and reuses relevant parameters across tasks to enable effective knowledge transfer while preventing interference.
%
We evaluate TSN-Affinity on two continual offline RL benchmarks covering both discrete and continuous control: a suite of classic Atari games and the multi-goal panda-gym environments utilizing the Franka Emika Panda robotic arm. Our experimental study shows that sparse task-specific subnetworks provide strong retention, while our proposed Affinity Routing further improves final multi-task performance, especially on the discrete Atari tasks. At the same time, the results on the Panda benchmark highlight the greater difficulty of knowledge transfer in heterogeneous continuous-control settings.

Our contributions can be summarized as follows: \\
1. TSN-Affinity, a novel CORL method that formulates knowledge transfer as similarity-driven routing over sparse task-specific subnetworks, dynamically selects and reuses parameters across tasks. \\
2. Affinity Routing, a CORL-specific mechanism that leverages action- and representation-level similarity to dynamically select and reuse parameters across tasks.\\
3. Two continuous offline reinforcement learning benchmarks covering visual discrete-control (Atari) and continuous-control (robotic manipulation) settings, enabling systematic analysis of retention and transfer. \\
4. An extensive experimental evaluation demonstrating that architectural approaches can completely mitigate forgetting and that similarity-driven reuse significantly improves multi-task performance.
}



\section{Related Work}
\label{sec:related}
\paragraph{Architectural continual learning.}
Architectural methods mitigate forgetting by allocating task-specific parameters, masks, or subnetworks within a shared backbone. Representative examples include PackNet, which iteratively prunes and reallocates weights, Piggyback, which learns binary task masks over a fixed network, and HAT, which uses task-specific hard attention to gate parameter usage \cite{mallya2018packnet,mallya2018piggyback,serra2018hat}. Our work is most closely related to the recent architectural CL line: Ada-QPackNet combines pruning and quantization for capacity-aware continual learning, while TinySubNets extends this idea with adaptive quantization, weight sharing, replay memory, and KL-based routing \cite{pietron2023adaqpacknet,pietron2025tsn}. We also build on the benchmark-oriented work, which stressed the need for more realistic and heterogeneous continual learning evaluation \cite{faber2024mnist2imagenet}.

\paragraph{Offline reinforcement learning and sequence modeling.}
Offline RL studies how to learn policies entirely from static datasets without additional environment interaction. The central challenge in this setting is the distributional shift between the training data and the states visited during online evaluation. This shift causes standard off-policy algorithms to query out-of-distribution (OOD) actions, resulting in overly optimistic value function estimates and policy collapse \cite{levine2020offline}. To mitigate this, many offline RL methods, such as CQL \cite{kumar2020cql} and SAC-N \cite{an2021uncertainty}, address the instability by enforcing conservative value updates. Alternatively, architectures like the Decision Transformer and Trajectory Transformer demonstrate that offline RL can be effectively formulated as a sequence modeling problem over trajectories, bypassing explicit value-function optimization entirely \cite{chen2021decisiontransformer,janner2021trajectorytransformer}. Because these models treat action generation as a supervised learning problem, they naturally constrain their predictions to the empirical distribution of the dataset, largely avoiding OOD actions. Our work adopts this sequence-modeling perspective via the Decision Transformer, extending it to a continual setting where tasks arrive sequentially.

\paragraph{Continual reinforcement learning and continual offline RL.}
Replay remains one of the strongest general baselines for continual adaptation in reinforcement learning \cite{rolnick2019experience}. More recently, replay-centric continual RL methods such as RECALL revisited the role of replay under policy drift and reward-scale mismatch \cite{zhang2023recall}. In continual offline RL, OER formulated sequential offline task learning with a bounded replay buffer and showed the effectiveness of offline replay under distribution mismatch \cite{gai2023oer}. Our contribution is complementary: instead of relying only on sample-level replay, we study parameter-level reuse through sparse task-specific subnetworks and control-aware routing.

\paragraph{Continual RL with transformer-based models.}
Published work on transformer-based continual RL remains limited. Foundational multi-task architectures, such as the Multi-Game Decision Transformer \cite{lee2022multi}, demonstrate that DTs can successfully represent diverse behaviors when past and present data are accumulated and trained jointly, effectively serving as a cumulative continual learning baseline. Moving beyond this joint-training setting to strict sequential task arrival, P2DT introduced a progressive prompt-based Decision Transformer to mitigate forgetting by appending task-specific decision tokens \cite{wang2024p2dt}. Similarly, L2M studied continual adaptation by modulating a frozen pre-trained model, reporting strong results on continuous-control benchmarks \cite{schmied2023l2m}. More recently, VQ-CD addressed continual offline RL with heterogeneous state and action spaces through quantized space alignment and selective weight activation in a diffusion-based framework \cite{hu2025vqcd}. In contrast, our work focuses on architectural continual learning for Decision Transformer backbones through sparse task-specific subnetworks and reuse-aware routing, evaluating this perspective on both visual discrete-control (Atari) and continuous robotic-manipulation (Panda) benchmarks.



\section{Methodology}
\label{sec:method}

\begin{figure}[t]
    \centering
    \includegraphics[width=\linewidth]{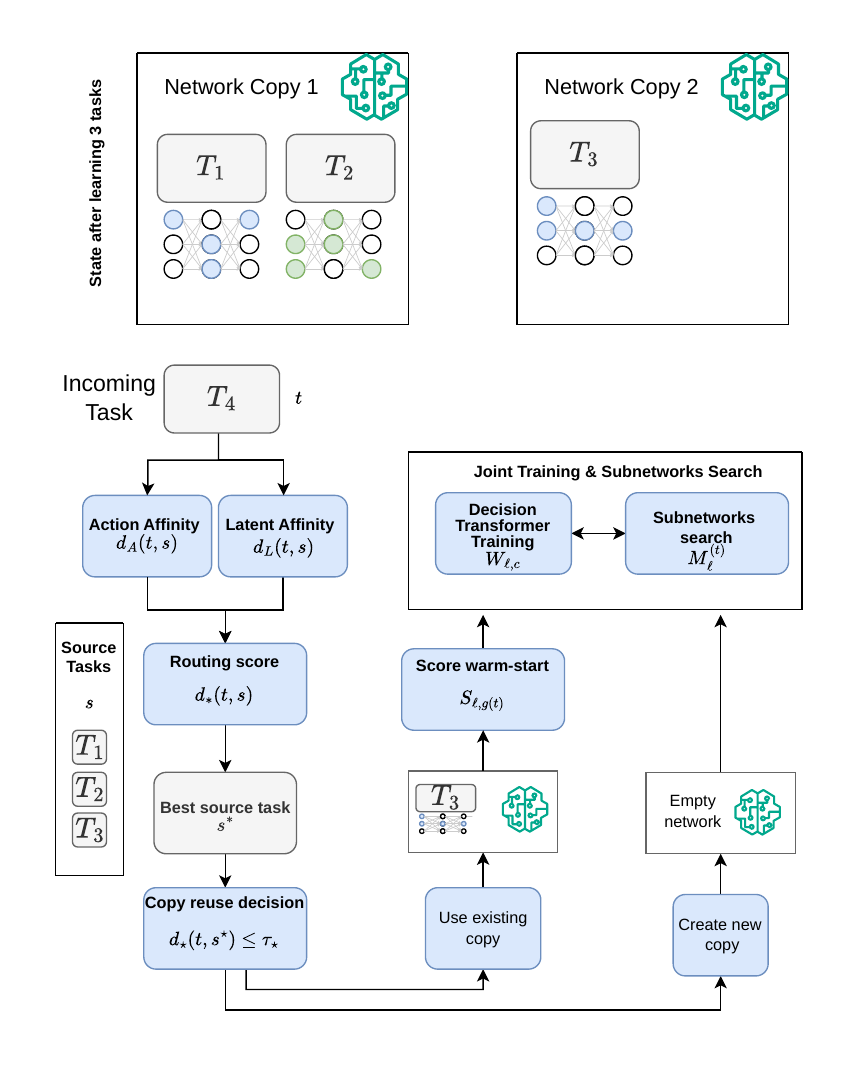}
    \caption{Overview of the proposed TSN-Affinity method.}
    \label{fig:tsn_affinity_overview}
\end{figure}

{\color{black}
We consider a CORL setting with a sequence of tasks
$\mathcal{T}_1,\mathcal{T}_2,\dots,\mathcal{T}_N$.
For each task $\mathcal{T}_t$, the learner receives a fixed offline dataset
$\mathcal{D}_t = \{\tau_i^{(t)}\}_{i=1}^{M_t}$ of trajectories.
Tasks are revealed sequentially, and the learner must acquire each new task while preserving performance on previously learned ones.
No environment interaction is used during model optimization.

Our main contribution is \textbf{TSN-Affinity}, a continual offline RL method that combines
(i) sparse task-specific subnetworks,
(ii) copy-level routing across multiple model instances,
and
(iii) RL-aware affinity scores for transfer-aware reuse.
Instead of relying solely on replay to preserve past behavior, TSN-Affinity stores previously learned task subnetworks and decides, for each incoming task, whether it should reuse an existing model copy or initialize a new one.
When reuse is selected, previously learned weights are not fine-tuned further; they remain frozen and are reused as fixed functional components, while the new task learns only its own sparse mask and any newly allocated trainable parameters.
Figure~\ref{fig:tsn_affinity_overview} illustrates this mechanism.

TSN-Affinity is built on a Decision Transformer (DT) backbone $f_\theta$.
For discrete-control domains such as Atari, the model predicts action logits and is trained with a cross-entropy objective over valid (non-padded) sequence positions.
For continuous-control domains such as Panda, the model predicts continuous actions and is trained with a masked mean-squared-error objective.
When Panda tasks have heterogeneous action dimensionalities, actions are padded to a shared global action dimension.
During training, an action-dimension mask is applied in the loss, while during evaluation the active output dimensionality is enforced through task-specific truncation.

We also study two reduced variants of the same framework:
\textbf{TSN-ReplayKL}, which replaces affinity routing by replay-memory similarity,
and
\textbf{TSN-Core}, which removes routing entirely and keeps a single sparse model copy.
These variants serve as controlled comparisons for isolating the contribution of affinity-based routing.

\subsection{Decision Transformer training objectives}
\label{subsec:dt_objectives}

Each trajectory is represented as a sequence of observations, actions, timesteps, and returns-to-go.
For a minibatch, let $\Omega$ denote the set of valid (non-padded) sequence positions. For a minibatch of size $B$ and context length $L$, $b \in \{1,\dots,B\}$ indexes the batch element, while $\ell \in \{1,\dots,L\}$ indexes the sequence position (time step) within that element. Let $\Omega \subseteq \{1,\dots,B\}\times\{1,\dots,L\}$ denote the set of valid (non-padded) sequence positions.

For Atari, the Decision Transformer outputs action logits $\hat{y}_{b,\ell}$, and the training objective for task $t$ is
\begin{equation}
\mathcal{L}_{\mathrm{Atari}}^{(t)}
=
\frac{1}{|\Omega|}
\sum_{(b,\ell)\in\Omega}
\mathrm{CE}\!\left(\hat{y}_{b,\ell}, a_{b,\ell}\right),
\end{equation}
where $CE$ is cross-entropy and the logits are computed from teacher-forced trajectory prefixes consisting of observations, past actions, returns-to-go, and timesteps.

For Panda, let $m_a^{(t)} \in \{0,1\}^{d_{\max}}$ denote the action dimension mask of task $t$, where $d_{\max}$ is the shared global action dimension.
The Panda objective is
\begin{equation}
\mathcal{L}_{\mathrm{Panda}}^{(t)}
=
\frac{1}{|\Omega|}
\sum_{(b,\ell)\in\Omega}
\frac{
\left\|
m_a^{(t)} \odot \big(\hat{a}_{b,\ell} - a_{b,\ell}\big)
\right\|_2^2
}{
\|m_a^{(t)}\|_1
},
\end{equation}
where $\hat{a}_{b,\ell}$ is the predicted continuous action.
Thus, the Panda loss is averaged only over the active action dimensions of the current task.

\subsection{Sparse task-specific subnetworks}
\label{subsec:tsn_parameterization}

TSN-Affinity operates on selected DT layers, denoted by $\mathcal{A}_{\mathrm{TSN}}$ 
In practice, these are linear and convolutional layers, and optionally, embedding layers.
Selected modules, such as the timestep embedding, may be excluded from conversion.
Each TSN-compatible layer $\ell \in \mathcal{A}_{\mathrm{TSN}}$ in model copy $c$ contains: i) a dense weight tensor $W_{\ell,c}$; ii) a trainable score tensor $S_{\ell,c}$ of the same shape, iii) and, when present, analogous tensors for the bias term.
For each task $t$, the method learns a binary task-specific mask
$M_{\ell}^{(t)} \in \{0,1\}^{\mathrm{shape}(W_{\ell,g(t)})}$,
where $g(t)$ denotes the selected model copy 
The mask is obtained by top-$k$ selection over the largest absolute scores among admissible parameters:
\begin{equation}
M_{\ell}^{(t)}
=
\operatorname{TopKMask}\!\left(
|S_{\ell,g(t)}|,
\rho_t,
F_{\ell,g(t)}^{(t-1)}
\right),
\end{equation}
where $\rho_t$ is the task keep ratio and $F_{\ell,g(t)}^{(t-1)}$ is the feasibility mask of parameters currently available to the new task.  
{Equivalently, if $F_{\ell,g(t)}^{(t-1)}$ denotes the feasibility mask, then the number of selected parameters is
\begin{equation}
k_{\ell}^{(t)}
=
\left\lceil
\rho_t \, \bigl\|F_{\ell,g(t)}^{(t-1)}\bigr\|_0
\right\rceil,
\end{equation}
so that, up to rounding,
\begin{equation}
\frac{\|M_{\ell}^{(t)}\|_0}{\|F_{\ell,g(t)}^{(t-1)}\|_0}
\approx
\rho_t.
\end{equation}
Thus, $\rho_t$ controls the density of the task-specific mask, whereas the corresponding sparsity level is $1-\rho_t$.}
If reuse is disabled, the task may select only previously unused weights.
If reuse is enabled, the task may also reactivate already occupied weights while keeping them frozen.
The effective weights used by task $t$ are therefore:
\begin{equation}
\widetilde{W}_{\ell}^{(t)}
=
M_{\ell}^{(t)} \odot W_{\ell,g(t)},
\end{equation}
where $\odot$ denotes element-wise multiplication.
Previously allocated weights are tracked by a copy-local occupancy mask:
\begin{equation}
O_{\ell,c}^{(t-1)}
=
\bigvee_{\substack{k<t\\g(k)=c}} M_{\ell}^{(k)},
\end{equation}
so occupancy, reuse, and protection are always handled locally within a selected copy.

\subsection{Training inside a selected copy}
\label{subsec:sparse_training}

Once a copy has been selected for task $t$, sparse training proceeds by jointly optimizing the DT objective and the score tensors that define the current task mask.
The task keep ratio $\rho_t$ can either be constant, where $\rho_t=\rho$ for all tasks, or follow an \texttt{equal\_remaining} schedule, where each task receives approximately an equal share of the remaining sparse capacity.

A key property of TSN-Affinity is that previously allocated weights remain protected.
Gradients on occupied parameters are masked as:
\begin{equation}
\nabla W_{\ell,g(t)}
\leftarrow
\big(1 - O_{\ell,g(t)}^{(t-1)}\big)\odot \nabla W_{\ell,g(t)}.
\end{equation}
In addition, after every optimizer step, protected parameters are restored to their pre-update values.
This makes weight protection strict even when optimizer state could otherwise introduce drift.

As a result, reuse does not mean re-training previously learned parameters.
Instead, it means that the new task may \emph{reactivate} them in its own sparse mask and use them as frozen functional components, while learning proceeds only through the current score tensors and any newly allocated trainable parameters.
Optionally, non-mask parameters outside the TSN-converted layers can also be frozen after the first task.

\subsection{TSN-Affinity: routing by RL-aware task affinity}
\label{subsec:tsn_affinity}

The core novelty of our method lies in the routing mechanism shown in Figure~\ref{fig:tsn_affinity_overview}.
Before training the incoming task $\mathcal{T}_t$, TSN-Affinity compares it to previously learned tasks and decides whether it should:
(i) reuse one of the existing model copies,
or
(ii) spawn a new copy.
Unlike replay-only routing, this decision is based on control-relevant affinity scores.

For the first task, no routing is required, and the method reduces to sparse TSN training in a single copy.
For later tasks, we consider three routing modes:
\textbf{TSN-Affinity-A} (action affinity),
\textbf{TSN-Affinity-L} (latent affinity),
and
\textbf{TSN-Affinity-H} (hybrid affinity).

\subsubsection{Action affinity}

Action affinity asks whether a previously learned subnetwork already explains expert behavior on the incoming task.
For each candidate source task $s$, we activate the source-task copy and source-task mask, then evaluate the new-task trajectories under teacher forcing.

For Atari, the action-affinity score is the average cross-entropy:
\begin{equation}
d_{\mathrm{A}}(t,s)
=
\frac{1}{|\Omega|}
\sum_{(b,\ell)\in\Omega}
\mathrm{CE}\!\left(
\hat{y}_{b,\ell}^{(s\rightarrow t)},
a_{b,\ell}
\right),
\end{equation}
where $\hat{y}_{b,\ell}^{(s\rightarrow t)}$ are the action logits produced by the DT with source-task mask $s$ on trajectories of task $t$.

For Panda, the action-affinity score is the masked mean-squared error:
\begin{equation}
d_{\mathrm{A}}(t,s)
=
\frac{1}{|\Omega|}
\sum_{(b,\ell)\in\Omega}
\frac{
\left\|
m_a^{(t)} \odot \big(\hat{a}_{b,\ell}^{(s\rightarrow t)} - a_{b,\ell}\big)
\right\|_2^2
}{
\|m_a^{(t)}\|_1
}.
\end{equation}

Small values of $d_{\mathrm{A}}(t,s)$ indicate that the source-task subnetwork already matches the demonstrated behavior of the new task.
This makes action affinity a direct measure of transfer at the policy level.

\subsubsection{Latent affinity}

Action compatibility alone does not capture whether two tasks are represented similarly inside the model.
TSN-Affinity therefore also compares tasks at the representation level.
For each learned task $s$, we store latent statistics computed from its task memory under its own copy and task mask.
The latent vectors are extracted from the DT observation encoder:
\begin{equation}
z = \phi_s(o),
\end{equation}
where $\phi_s$ denotes the observation encoder with source-task mask $s$ activated.
For each task memory, we fit a diagonal Gaussian approximation:
\begin{equation}
z \sim \mathcal{N}(\mu,\Sigma),
\qquad
\Sigma = \mathrm{diag}(\sigma_1^2,\dots,\sigma_H^2).
\end{equation}

To evaluate a new task $t$ against source task $s$, we pass the new-task memory through the \emph{same source-task encoder $\mathcal{N}$} and compare the resulting latent statistics to those previously stored for $s$:
\begin{equation}
d_{\mathrm{L}}(t,s)
=
\frac{1}{2}
\Big(
\mathrm{KL}(\mathcal{N}_{t\rightarrow s}\|\mathcal{N}_{s})
+
\mathrm{KL}(\mathcal{N}_{s}\|\mathcal{N}_{t\rightarrow s})
\Big).
\end{equation}


{We use the symmetric KL divergence rather than a one-way KL term, because standard KL is asymmetric and would otherwise make the latent-affinity score depend on which task distribution is treated as the reference.}
This score measures whether the incoming task induces representations similar to those already supported by a given source copy.

\subsubsection{Hybrid affinity}

The hybrid mode combines both signals:
\begin{equation}
d_{\mathrm{H}}(t,s)
=
\alpha\,\widetilde{d}_{\mathrm{A}}(t,s)
+
(1-\alpha)\,\widetilde{d}_{\mathrm{L}}(t,s),
\end{equation}
where $\alpha\in[0,1]$ controls the trade-off.
When multiple candidate source tasks are available, the two score families are normalized across candidates before combination.
When only one candidate exists, the direct scores are used without cross-task normalization.

\subsubsection{Routing decision and copy reuse}

Let $d_\star(t,s)$ denote the active routing criterion, where $\star \in \{\mathrm{A},\mathrm{L},\mathrm{H}\}$.
The best source task is selected as:
\begin{equation}
s^\star
=
\arg\min_{s<t} d_\star(t,s).
\end{equation}
If the following condition applies:
\begin{equation}
d_\star(t,s^\star) \le \tau_\star,
\end{equation}
the new task is assigned to the copy of task $s^\star$.
Otherwise, a new copy is created.
If a maximum number of copies is imposed and already reached, the task falls back to the best existing copy.
{Thus, $\tau_\star$ acts as a reuse-capacity control: lower values make reuse more conservative and tend to create more copies, whereas higher values force more tasks into existing copies.}

The resulting behavior is exactly the one depicted in Figure~\ref{fig:tsn_affinity_overview}: the method either reuses a selected copy, where previously learned components remain frozen and protected while new sparse trainable connections are added, or spawns a fresh copy for the incoming task.

{When a source task is selected, the new task may optionally warm-start mask scores from the source mask; details are given in Appendix~\ref{app:score_warm_start}.}

\subsection{Reduced comparison variants}
\label{subsec:reduced_variants}

To isolate the contribution of affinity-based routing, we also study two reduced variants of the same framework.

\paragraph{TSN-ReplayKL.}
TSN-ReplayKL keeps the multi-copy TSN structure but replaces RL-aware routing with a replay-memory similarity heuristic.
{For each task $\mathcal{T}_t$, it constructs a compact observation memory $\mathcal{M}_t=\{x_1^{(t)},\dots,x_K^{(t)}\}$. Each flattened memory sample is converted into a pseudo-probability vector using a softmax transform, and the routing score is:
\begin{equation}
d_{\mathrm{RKL}}(t,s)
=
\frac{1}{K'}
\sum_{i=1}^{K'}
\mathrm{KL}\!\left(
p_i^{(s)} \,\|\, p_i^{(t)}
\right),
\end{equation}
where $p_i^{(t)}=\mathrm{softmax}(x_i^{(t)})$ and $K'=\min(|\mathcal{M}_t|,|\mathcal{M}_s|)$. Thus, TSN-ReplayKL retains copy-level reuse but uses observation-level similarity rather than control-aware affinity.}

\paragraph{TSN-Core.}
TSN-Core removes copy routing altogether and keeps a single evolving model copy.
Each task learns a fresh sparse subnetwork within that copy, while previously occupied parameters remain protected.
TSN-Core, therefore, isolates the effect of sparse task-specific parameterization without any routing or copy selection.

\subsection{Task-specific inference}
\label{subsec:task_specific_inference}

All variants are evaluated in the task-incremental continual learning protocol.
At test time, task identity is provided and used to activate the corresponding task-specific mask.
For copy-based variants, this also selects the corresponding model copy.
For Panda reuse variants, task-specific observation normalization statistics and the active action dimensionality are restored together with the selected task mask.
}



\section{Benchmarks}
\label{sec:benchmarks}

We evaluate CORL on two complementary benchmark families. The first benchmark targets high-dimensional visual decision making with discrete actions, while the second focuses on low-dimensional continuous-control manipulation. Together, they test whether subnetwork reuse remains effective across different observation modalities, action spaces, horizon lengths, and transfer regimes. 
Additional experimental details are reported in Appendix \ref{app:benchmarks}.

\subsection{Atari Continual Decision Making}
\label{subsec:atari_benchmark}

Our first benchmark is a continual offline Atari suite constructed from five ALE tasks:
\texttt{Breakout}, \texttt{Alien}, \texttt{Atlantis}, \texttt{Boxing}, and \texttt{Centipede}. This benchmark operates on high-dimensional visual observations and discrete action spaces, making it substantially different from standard low-dimensional control settings.
%
%
%
In the experiments discussed in this work, we use the following task order:
\emph{Breakout} $\rightarrow$ \emph{Alien} $\rightarrow$ \emph{Atlantis}
$\rightarrow$ \emph{Boxing} $\rightarrow$ \emph{Centipede}.
{Evaluation details, horizons, and return-conditioning settings are reported in Appendix~\ref{app:benchmarks}.}

\subsection{Panda Continual Manipulation}
\label{subsec:panda_benchmark}

Our second benchmark utilizes the \texttt{panda-gym} simulation suite, which models the continuous control of a Franka Emika Panda robotic arm. From this framework, we construct a compact continual-learning suite using three dense-reward manipulation tasks: \texttt{PandaReachDense-v3}, \texttt{PandaPushDense-v3}, and \texttt{PandaPickAndPlaceDense-v3}. These tasks share an underlying kinematic structure but progressively increase in difficulty. Specifically, Reach is a short-horizon end-effector positioning problem, Push demands controlled interaction with an unattached object, and Pick-and-Place further introduces grasping mechanics and vertical object translation.
%
%
%
%
Unless stated otherwise, the Panda task order is
$\texttt{PandaReach} \rightarrow \texttt{PandaPush} \rightarrow \texttt{PandaPickAndPlace}.$
{Evaluation details, horizons, and return-conditioning settings are reported in Appendix~\ref{app:benchmarks}.}

\subsection{Evaluation Protocol}
\label{subsec:eval_protocol}

Let $T$ denote the number of tasks in a continual sequence. After training on task $i$, each method is evaluated on all tasks, yielding a performance matrix
$\mathbf{P} \in \mathbb{R}^{T \times T},$
where $P_{i,j}$ is the average return on task $j$ after learning task $i$.
We report the full performance matrix for completeness, but continual-learning metrics are computed from the task-incremental part of the matrix, i.e., from entries with $j \le i$. For task-specific architectural methods, task identity is provided at evaluation time in order to activate the corresponding task-specific subnetwork, following the standard task-incremental continual learning protocol.
In addition to aggregate metrics, we focus on:
(i) task-wise final performance,
(ii) preservation of early tasks after later updates,
and
(iii) the ability of reuse-based methods to improve transfer without collapsing previously learned behavior.
This protocol is identical across Atari and Panda, allowing us to compare the behavior of the same reuse mechanism under visual discrete-control and continuous-control regimes.

\section{Results}
\label{sec:results}

{The routing threshold in reuse-based TSN variants is a capacity-control hyperparameter: it determines when an incoming task is similar enough to reuse an existing copy rather than allocate a new one. For each benchmark, we use a small grid of threshold values and keep the selected threshold fixed across the full task sequence. The full grid, together with single-task reference scores and auxiliary runs, is reported in Appendix~\ref{app:additional_tables}, so the performance--capacity trade-off induced by this choice is explicit.}

\subsection{Atari benchmark}
\label{subsec:results-atari}

We evaluate all methods on a five-task continual offline Atari benchmark with the task order
\emph{Breakout} $\rightarrow$ \emph{Alien} $\rightarrow$ \emph{Atlantis} $\rightarrow$ \emph{Boxing} $\rightarrow$ \emph{Centipede}.
{Because the games operate on very different score scales, we summarize final performance with the expert-normalized mean return:
\begin{equation}
\mathrm{NormAvg}
=
\frac{1}{N}
\sum_{j=1}^{N}
100 \cdot \frac{P_{N,j}}{R_j^\star},
\label{eq:atari_normavg}
\end{equation}
where $P_{N,j}$ is the final score on task $j$ after learning all $N$ tasks and $R_j^\star$ is the target return used for Decision Transformer conditioning.
We also measure forgetting by
\begin{equation}
F_j = \max_{t \in \{j,\dots,N\}} P_{t,j} - P_{N,j},
\label{eq:atari_forgetting}
\end{equation}
where larger values indicate stronger degradation after a task has already been learned.}
Single-task Atari reference scores are reported in Appendix~\ref{tab:app_atari_single_reference}.

\begin{table*}[t]
\centering
\small
\setlength{\tabcolsep}{4pt}
\begin{tabular}{lccccccc}
\hline
Method & Copies & Breakout & Alien & Atlantis & Boxing & Centipede & NormAvg (\%) \\
\hline
Naive & 1 & 0 & 27 & 10 & -10 & 92 & 18.0 \\
Cumulative & 1 & 57 & 52 & 1492 & 92 & 108 & 76.2 \\
EWC & 1 & 0 & 29 & 7 & -3 & 75 & 17.3 \\
SI & 1 & 0 & 20 & 23 & -1 & 68 & 14.8 \\
\hline
TSN-Core & 1 & \textbf{114} & 52 & \textbf{1564} & \textbf{97} & 103 & 87.5 \\
TSN-ReplayKL & 1 & 103 & 52 & 1522 & 90 & \textbf{131} & 87.7 \\
TSN-Affinity-A ($\tau_A=10$) & 4 & 97 & \textbf{73} & 1511 & 92 & 117 & \textbf{89.7} \\
TSN-Affinity-L ($\tau_L=35$) & 5 & 97 & \textbf{73} & 1501 & 92 & 117 & 89.6 \\
TSN-Affinity-H ($\tau_H=0.50,\alpha=0.70$) & 1 & 97 & 38 & 1511 & 92 & 62 & 73.4 \\
\hline
\end{tabular}
\caption{{Final Atari results for the selected operating point of each method family on the Breakout-first continual offline RL benchmark. ``Copies'' denotes the number of model copies created by the reuse mechanism and serves as a coarse proxy for additional capacity and memory overhead. The full threshold grid is reported in Appendix~\ref{app:additional_tables}.}}
\label{tab:atari_main_results}
\end{table*}

Table~\ref{tab:atari_main_results} shows three main trends.

First, the dense baselines are clearly weaker in this benchmark.
Naive, EWC, and SI all remain in the same low-performance regime, with normalized final returns between $14.8\%$ and $18.0\%$, which indicates severe catastrophic forgetting.
Among the dense methods, cumulative replay is clearly the strongest baseline.
It improves substantially over Naive, EWC, and SI and reaches $76.2\%$ normalized performance, but it still remains well below the best TSN-based methods.
This shows that replay alone is helpful, yet it does not remove backward interference when all tasks continue to share a single dense parameterization.

Second, sparse task-specific subnetworks provide an exceptionally strong retention regime in Atari.
In all reported TSN-based Atari runs, final forgetting is zero.
Once a task-specific sparse subnetwork has been allocated, subsequent tasks do not overwrite it.
As a result, the main challenge in Atari is not forgetting itself, but rather how efficiently the model allocates and reuses capacity for future tasks.
{Capacity, routing-memory, and quantization settings are summarized in Appendix~\ref{app:experimental_setup}.}

Third, routing and reuse improve final multi-task performance on top of this already strong retention.
TSN-ReplayKL improves over TSN-Core, and the affinity-based variants improve further.
At the selected operating point, TSN-Affinity-A with $\tau_A=10$ reaches $89.7\%$ normalized performance, closely followed by TSN-Affinity-L with $\tau_L=35$ at $89.6\%$.
{Because the best-performing configurations use additional model copies, these gains should be interpreted as a performance--capacity trade-off; details are provided in Appendix~\ref{app:capacity_discussion}.}
{The effect of the routing threshold is analyzed explicitly in Appendix~\ref{app:additional_tables}; the same threshold is fixed across the whole Atari task sequence for each reported run.}
Both TSN-Affinity-A and TSN-Affinity-L are also competitive with the single-task Atari references in Appendix~\ref{tab:app_atari_single_reference}, which indicates that the best continual TSN variants operate very close to the performance range of individually trained models.

Overall, the Atari experiments support three conclusions:
i) dense continual baselines, including replay and regularization methods, remain substantially weaker than sparse TSN-based models,
ii) sparse task-specific subnetworks are already sufficient to eliminate forgetting in the reported Atari setting, and
iii) RL-aware affinity routing improves final multi-task performance further, with the strongest results obtained by action-based and latent-based routing at the cost of additional model copies.

Appendix~\ref{app:additional_tables} contains the single-task reference tables, the full sets of tested threshold values, and auxiliary repeated runs extracted from the available logs.

\subsection{Panda benchmark}
\label{subsec:results-panda}

We evaluate all methods on a three-task continual offline Panda benchmark with the task order
\emph{PandaReach} $\rightarrow$ \emph{PandaPush} $\rightarrow$ \emph{PandaPickAndPlace}.
Because the target returns in Panda are very close to zero, expert-normalized percentages are not stable or particularly interpretable.
{We therefore report final task returns directly and summarize overall quality using the mean absolute deviation from the target return:
\begin{equation}
\mathrm{AvgGap}
=
\frac{1}{N}
\sum_{j=1}^{N}
\left|P_{N,j} - R_j^\star\right|.
\label{eq:panda_avggap}
\end{equation}
We also report the mean forgetting, denoted by $\mathrm{AvgF}$, obtained by averaging the standard forgetting score from Eq.~\eqref{eq:atari_forgetting} over previously learned tasks.}
Higher returns (i.e., values closer to zero) are better, while lower $\mathrm{AvgGap}$ and $\mathrm{AvgF}$ are preferable.
Single-task Panda reference scores are reported in Appendix~\ref{tab:app_panda_single_reference}.

\begin{table*}[ht]
\centering
\small
\setlength{\tabcolsep}{2pt}
\begin{tabular}{lcccccc}
\hline
Method & Copies & PandaReach $\uparrow$ & PandaPush $\uparrow$ & PandaPickAndPlace $\uparrow$ & AvgGap $\downarrow$ & AvgF $\downarrow$ \\
\hline
Naive & 1 & -10.902 & -8.195 & -1.311 & 6.657 & 8.977 \\
Cumulative & 1 & -0.189 & -1.120 & -1.314 & 0.729 & 0.071 \\
EWC & 1 & -13.863 & -7.979 & -1.490 & 7.632 & 10.352 \\
SI & 1 & -1.754 & -7.011 & -11.890 & 6.739 & 3.012 \\
\hline
TSN-Core & 1 & -0.235 & -2.589 & -1.427 & 1.271 & 0.857 \\
TSN-ReplayKL & 2 & -7.865 & -1.026 & -2.061 & 3.505 & 3.838 \\
TSN-Affinity-A ($\tau_A=0.45$) & 3 & -0.189 & -0.972 & -1.422 & 0.715 & 0.001 \\
TSN-Affinity-L ($\tau_L=25$) & 3 & -0.189 & -0.949 & -1.308 & \textbf{0.670} & \textbf{0.000} \\
TSN-Affinity-H ($\tau_H=0.50,\alpha=0.70$) & 2 & -1.482 & -0.949 & -2.643 & 1.546 & 0.647 \\
\hline
\end{tabular}
\caption{{Final Panda results for the selected operating point of each method family. ``Copies'' denotes the number of model copies created by the reuse mechanism and serves as a coarse proxy for additional capacity and memory overhead. Auxiliary runs and threshold settings are reported in Appendix~\ref{app:additional_tables}.}}
\label{tab:panda_main_results}
\end{table*}

Table~\ref{tab:panda_main_results} shows that Panda is substantially more challenging than Atari from the continual transfer perspective.
Unlike Atari, Panda does not produce a universal zero-forgetting regime across the whole TSN family.
Instead, the methods must balance retention against transfer to later tasks.

First, the dense baselines are generally weak.
Naive and EWC suffer from severe forgetting, with large $\mathrm{AvgF}$ values and poor final performance on \emph{PandaReach} and \emph{PandaPush}.
SI is somewhat more stable than EWC but still forgets substantially and collapses on \emph{PandaPickAndPlace}.
Among the dense baselines, cumulative replay is clearly the strongest: it achieves the best dense-model trade-off, with $\mathrm{AvgGap}=0.729$ and near-optimal retention on \emph{PandaReach}.

Second, the Panda benchmark highlights the limitations of replay-memory KL routing in heterogeneous continuous control.
TSN-Core improves considerably over Naive, EWC, and SI, but it does not match cumulative replay.
TSN-ReplayKL is worse still, especially on \emph{PandaReach}, which suggests that observation-memory KL is not a reliable routing signal in this setting.
{A detailed discussion of this behavior and the replay-memory comparison is provided in Appendix~\ref{app:capacity_discussion}.}

Third, the Panda results should be interpreted together with their capacity cost.
At the selected operating point, TSN-Affinity-L with $\tau_L=25$ reaches the lowest $\mathrm{AvgGap}$ of $0.670$ and zero forgetting, but uses 3 model copies.
Relative to TSN-Core, this yields a clear improvement in both transfer and retention.
However, relative to cumulative replay, the gain is more modest, improving $\mathrm{AvgGap}$ only from $0.729$ to $0.670$.
{Hence, the Panda results also reflect a performance--capacity trade-off rather than a free improvement.}
{The effect of the routing threshold is analyzed explicitly in Appendix~\ref{section:thresholds_analysis}; the same threshold is fixed across the whole Panda task sequence for each reported run.}
{Capacity, routing-memory, and quantization settings are summarized in Appendix~\ref{app:experimental_setup}.}

Finally, the single-task Panda results in Appendix~\ref{tab:app_panda_single_reference} show that Panda remains difficult even without continual interference.
In particular, \emph{PandaPush} and \emph{PandaPickAndPlace} do not approach zero return in the single-task setting either.
Thus, the Panda gap is not caused by forgetting alone; it also reflects the intrinsic difficulty of offline continuous-control transfer in a heterogeneous action space.

Overall, the Panda experiments support three conclusions:
i) Panda is a substantially harder continual offline RL benchmark than Atari,
ii) replay-memory KL routing is not reliable enough in this heterogeneous continuous-control setting, and
iii) RL-aware affinity routing provides the best trade-off between transfer and retention.

Appendix~\ref{app:additional_tables} contains the single-task reference tables, the full sets of tested threshold values, and auxiliary repeated runs extracted from the available logs.

\section{Conclusion}
In this work, we investigated the challenges of CORL, where agents must learn a sequence of tasks from previously collected datasets without interacting with the environment. We proposed TSN-Affinity, a novel architectural approach that combines sparse task-specific subnetworks with similarity-driven parameter reuse. By formulating knowledge transfer as a routing problem based on action and latent affinities, our method effectively decouples the learning of new tasks from the preservation of previously acquired knowledge.

Our evaluation on the Atari and Panda benchmarks shows that architectural continual learning can provide strong retention and robust transfer in settings where replay-based baselines struggle with memory overhead or distribution mismatch. Specifically, the affinity-based routing mechanisms consistently outperformed replay-memory heuristics, particularly in the heterogeneous continuous-control domain of Panda. 
The results indicate that direct action and latent similarity are reliable indicators of routing decisions. 

Future work will focus on 
investigating the generalization of affinity scores to larger, more diverse benchmarks. 
Another interesting direction is extending the affinity routing framework to online continual learning settings, which could further elucidate the role of architectural reuse when environment interactions are feasible.
Finally, we will extend TinySubnetworks with hybrid pruning and adaptive quantization strategies.










\clearpage
\bibliographystyle{siamplain}

\clearpage
\appendix
\onecolumn

\section{Benchmarks}
\label{app:benchmarks}
In both benchmarks, tasks are revealed sequentially: at stage $t$, the learner receives only the offline dataset for the current task, while access to previous data is available only if a method explicitly stores or reuses it.

\subsection{Atari Continual Decision Making}


We use expert offline trajectories stored in \texttt{expert\_minari\_dqn.npz} format. The environment pipeline is matched to the dataset through a Minari-like DQN preprocessing setup with frame stacking of $4$, frame skip of $4$, and $84 \times 84$ observations. We additionally perform a replay-consistency check by replaying expert actions in the environment and verifying that the obtained return matches the dataset return, which ensures that the simulator and offline data are aligned.

The Decision Transformer uses a shared discrete-action interface over the whole task sequence. Observations have shape $(4,84,84)$ after preprocessing, while the action vocabulary is shared across tasks using the maximum action-space size observed in the benchmark.
{Evaluation is performed for $20$ episodes per task using greedy action selection. The maximum evaluation horizon is $27{,}000$ environment steps. Return-conditioning targets are derived independently for each task from the corresponding offline expert data.}


\subsection{Panda Continual Manipulation}
\label{app:panda_benchmark}



For each task, we use an offline expert dataset. Observations are formed by concatenating \texttt{observation}, \texttt{achieved\_goal}, and \texttt{desired\_goal}. In our setup, the dataset representation additionally includes a normalized time feature, which is matched during environment interaction through a dedicated wrapper.

Because the tasks have different raw observation and action dimensions, we map them into a shared continual-learning interface by padding all observation and action vectors to global dimensions. In practice, shorter vectors are zero-padded in the trailing coordinates, while Panda action losses are computed only over the task-active action dimensions. During evaluation, the model output is truncated back to the original task-specific action dimensionality. In the experiments reported here, the shared dimensions are $26$ for observations and $4$ for actions, allowing all methods to use a single Decision Transformer backbone across the whole Panda sequence.
{Evaluation is performed for $20$ episodes per task, with a maximum horizon of $50$ environment steps. We report the average episodic return. Return-conditioning uses fixed task-specific targets corresponding to expert-level returns in our setup.}

Single-task Panda reference scores are reported in Appendix~\ref{tab:app_panda_single_reference}.


\section{Experimental setup}
\label{app:experimental_setup}
{
For the TSN-based methods, we use a routing memory of 256 samples per task. This memory is used only for routing and latent-statistics estimation and is not replayed for gradient-based optimization. All Atari TSN experiments use a constant keep ratio of $\rho=0.5$, whereas all Panda TSN experiments use $\rho=0.33$, corresponding to task-wise sparsity levels of 50\% and 67\%, respectively, in TSN-converted layers.  We used benchmark-specific keep ratios chosen empirically: $\rho=0.5$ for Atari and $\rho=0.33$ for Panda, corresponding to task-wise sparsity levels of 50\% and 67\%, respectively, in TSN-converted layers. The larger Atari value reflects a denser per-task allocation in the more homogeneous discrete-control setting, whereas the smaller Panda value yields a more conservative allocation in the more heterogeneous continuous-control setting. Since the keep ratio is applied to the currently feasible parameter set, these values correspond to approximate single-copy occupancies of $1-(1-\rho)^5 \approx 96.9\%$ in Atari and $1-(1-\rho)^3 \approx 69.9\%$ in Panda when reuse is disabled. Quantization was disabled in all reported experiments in order to isolate the effect of sparse masking and routing. For Panda, the cumulative baseline uses a rehearsal capacity of 5000, while in Atari cumulative replay is controlled by a replay mix of 0.5 and the runner does not expose a separate replay-capacity limit.
}
\begin{table}[!htbp]
\centering
\setlength{\tabcolsep}{5pt}
\begin{tabular}{lccccc}
\hline
Benchmark & $\rho$ & Sparsity & Routing memory & Replay setting & Quantization \\
\hline
Atari & 0.50 & 50\% & 256 & mix = 0.5 & off \\
Panda & 0.33 & 67\% & 256 & capacity = 5000 & off \\
\hline
\end{tabular}
\caption{Key benchmark-specific TSN and replay settings used in the reported experiments.}
\label{tab:app_key_settings}
\end{table}

\section{Additional Method and Metric Details}
\label{app:method_metric_details}

\subsection{TSN Mask Density}
\label{app:tsn_mask_details}
{The mask-density equations are included in the main Methodology section. In the reported experiments, we use constant keep ratios of $\rho=0.5$ for Atari and $\rho=0.33$ for Panda, corresponding to task-wise sparsity levels of 50\% and 67\%, respectively, in TSN-converted layers.}

\subsection{Score Warm-Start}
\label{app:score_warm_start}
{
Once a source task $s^\star$ has been selected, the score tensors of the new task can optionally be warm-started from the source-task mask:
\begin{equation}
S_{\ell,g(t)}
\leftarrow
\lambda M_{\ell}^{(s^\star)} + \epsilon,
\end{equation}
where $\lambda > 0$ controls the warm-start strength, and $\epsilon$ is a small random noise term.
This warm-start does not copy or fine-tune old weights; it only biases sparse mask search toward structures that were already useful for the selected source task.
}



\subsection{Capacity and Replay-Memory Discussion}
\label{app:capacity_discussion}
{
For Atari, the memory used by TSN-ReplayKL and TSN-Affinity is a routing memory of 256 samples per task rather than a rehearsal buffer. Thus, the gap between TSN-ReplayKL and Cumulative is not caused by larger replay storage, but by the fact that TSN methods reduce interference through task-specific sparse allocation and copy-level routing, whereas Cumulative continues to optimize a single dense model on mixed past-task data.
The best Atari configurations use 4 and 5 model copies for TSN-Affinity-A and TSN-Affinity-L, respectively, whereas TSN-Core and TSN-ReplayKL use only a single copy in the reported setting. The roughly two-point gain over the strongest single-copy TSN baselines should therefore be interpreted as a performance--capacity trade-off rather than a free improvement.

In Panda, TSN-ReplayKL can underperform even the single-copy TSN-Core baseline because its routing decision is based on observation-level replay-memory similarity rather than control-aware compatibility. When KL-based routing creates an additional copy, the new copy is initialized from scratch rather than inheriting previously trained dense weights, so the method may lose beneficial sharing without obtaining strong transfer in return.
This also explains the large gap to cumulative replay: cumulative replay repeatedly optimizes on real past-task data, whereas TSN-ReplayKL uses only a small routing memory for copy selection and does not revisit past datasets in the same way during optimization. In our setup, the Panda cumulative baseline uses a rehearsal capacity of 5000, whereas TSN-ReplayKL and TSN-Affinity use routing memories of 256 samples per task.
{The selected Panda operating point uses 3 model copies, so this result also reflects a performance--capacity trade-off. More conservative reuse often improves performance by reducing forced reuse across heterogeneous tasks.} Since quantization was disabled in all reported Panda experiments, we do not report a compressed memory footprint here and instead use the number of copies as the main coarse proxy for additional capacity.
}

\section{Additional Tables}
\label{app:additional_tables}
The tables in this appendix complement the main results with three additional views.
First, we report single-task reference scores to separate continual-learning effects from the intrinsic difficulty of the underlying offline RL tasks.
Second, we provide threshold grids for the reuse-based TSN variants to show how the routing-capacity operating point affects final performance.
Third, we include auxiliary repeated runs to provide a compact view of run-to-run stability, especially in the more variable Panda setting.
Together, these tables provide additional support for the main conclusions of the paper while keeping the main text focused on the strongest representative configurations.

\begin{table}[!htbp]
\centering
\setlength{\tabcolsep}{5pt}
\begin{tabular}{lcc}
\hline
Task & Target $R^\star$ & Single-task DT \\
\hline
Breakout   & 114  & 112 \\
Alien      & 86   & 62 \\
Atlantis   & 1556 & 1505 \\
Boxing     & 97   & 86 \\
Centipede  & 135  & 117 \\
\hline
NormAvg (\%) & \multicolumn{2}{c}{88.5} \\
\hline
\end{tabular}
\caption{Single-task Atari references. Breakout and Atlantis are already strong, while Alien remains the hardest game.}
\label{tab:app_atari_single_reference}
\end{table}

\begin{table}[!htbp]
\centering
\setlength{\tabcolsep}{4pt}
\begin{tabular}{lccc}
\hline
Task & CL target $R^\star$ & Single-task at $R^\star$ & Best sweep \\
\hline
Reach        & -0.000320 & -0.243 & -0.243 \\
Push         & -0.436000 & -0.971 & -0.937 \\
Pick\&Place  & -0.001000 & -1.027 & -1.024 \\
\hline
\end{tabular}
\caption{Single-task Panda references. Push and Pick\&Place remain difficult even without continual interference.}
\label{tab:app_panda_single_reference}
\end{table}


\begin{table}[!htbp]
\centering
\setlength{\tabcolsep}{4pt}
\begin{tabular}{lccr}
\hline
Method/configuration & Run & NormAvg (\%) & AvgF \\
\hline
Cumulative & s0 & 67.2 & 359.0 \\
Cumulative & s1 & 76.2 & 20.3 \\
Cumulative & s2 & 62.6 & 8.5 \\
\hline
TSN-Core & s1 & 87.5 & 0.0 \\
TSN-Core & s2 & 81.5 & 0.0 \\
\hline
TSN-ReplayKL & s1 & 86.1 & 0.0 \\
TSN-ReplayKL & s2 & 87.7 & 0.0 \\
\hline
TSN-Affinity-A ($\tau_A=10$) & s0 & 89.7 & 0.0 \\
TSN-Affinity-A ($\tau_A=10$) & s1 & 89.2 & 0.0 \\
TSN-Affinity-A ($\tau_A=10$) & s2 & 82.2 & 0.0 \\
\hline
TSN-Affinity-L ($\tau_L=35$) & s0 & 89.6 & 0.0 \\
TSN-Affinity-L ($\tau_L=35$) & s1 & 86.9 & 0.0 \\
TSN-Affinity-L ($\tau_L=35$) & s2 & 81.6 & 0.0 \\
\hline
\end{tabular}
\caption{Auxiliary Atari runs. All listed TSN runs retain zero forgetting, whereas cumulative replay is more variable.}
\label{tab:app_atari_aux_runs}
\end{table}


\FloatBarrier

\subsection{Affinity Routing Thresholds Analysis}
\label{section:thresholds_analysis}

{Table \ref{tab:app_atari_thresholds} reports the threshold grid for affinity routing on the Atari benchmark.} 
For action-based routing, decreasing the threshold from $\tau_A=16$ to $\tau_A=10$ increases the number of copies from $2$ to $4$ and improves normalized performance from $88.5\%$ to $89.7\%$, mainly because \emph{Centipede} improves from $102$ to $117$.
At the same time, more aggressive reuse at $\tau_A=12$ or $\tau_A=14$ is clearly worse, dropping performance to about $81.9\%$.
{For latent routing, the strongest operating point in this grid is obtained at $\tau_L=35$.}
Increasing the threshold to $\tau_L=75$ reduces the number of copies from $5$ to $3$ and lowers performance from $89.6\%$ to $86.6\%$.
Intermediate values $\tau_L=45$ and $\tau_L=55$ are worse still, at about $82.3\%$.
Thus, in Atari, a more conservative reuse policy is preferable to forcing more tasks into the same copy.

\begin{table}[H]
\centering
\setlength{\tabcolsep}{3pt}
\resizebox{\textwidth}{!}{%
\begin{tabular}{lccccccc}
\hline
Variant & Copies & Breakout & Alien & Atlantis & Boxing & Centipede & NormAvg (\%) \\
\hline
TSN-Affinity-A ($\tau_A=10$) & 4 & 97 & 73 & 1511 & 92 & 117 & 89.7 \\
TSN-Affinity-A ($\tau_A=12$) & 2 & 97 & 73 & 1512 & 90 & 67 & 81.9 \\
TSN-Affinity-A ($\tau_A=14$) & 2 & 97 & 73 & 1512 & 90 & 67 & 81.9 \\
TSN-Affinity-A ($\tau_A=16$) & 2 & 97 & 73 & 1511 & 97 & 102 & 88.5 \\
\hline
TSN-Affinity-L ($\tau_L=35$) & 5 & 97 & 73 & 1501 & 92 & 117 & 89.6 \\
TSN-Affinity-L ($\tau_L=45$) & 3 & 97 & 73 & 1509 & 90 & 70 & 82.3 \\
TSN-Affinity-L ($\tau_L=55$) & 3 & 97 & 73 & 1509 & 90 & 70 & 82.3 \\
TSN-Affinity-L ($\tau_L=75$) & 3 & 97 & 73 & 1501 & 97 & 90 & 86.6 \\
\hline
TSN-Affinity-H ($\tau_H=0.50,\alpha=0.70$) & 1 & 97 & 38 & 1511 & 92 & 62 & 73.4 \\
TSN-Affinity-H ($\tau_H=0.50,\alpha=0.50$) & 1 & 97 & 38 & 1511 & 92 & 62 & 73.4 \\
TSN-Affinity-H ($\tau_H=0.55,\alpha=0.80$) & 1 & 97 & 38 & 1511 & 92 & 62 & 73.4 \\
\hline
\end{tabular}%
}
\caption{{Full Atari threshold grid. More conservative reuse often improves final performance in this Atari sequence, at the cost of additional copies, while the tested hybrid settings collapse to the same result.}}
\label{tab:app_atari_thresholds}
\end{table}

{Table \ref{tab:app_panda_aux_runs} reports the threshold grid and auxiliary reuse runs for affinity routing on the Panda benchmark.}
For action-based routing, we evaluated $\tau_A \in \{0.35,0.45,0.65\}$.
Reducing the threshold from $0.65$ to $0.45$ improves $\mathrm{AvgGap}$ from $0.737$ to $0.715$.
This gain comes mainly from better \emph{PandaPickAndPlace} performance, which improves from $-1.505$ to $-1.422$, while \emph{PandaPush} becomes slightly worse, from $-0.956$ to $-0.972$.
Lowering the threshold further from $0.45$ to $0.35$ does not change the final result, which indicates that the routing decisions are already saturated in that range.
{For latent routing, the strongest operating point in this grid is obtained at $\tau_L=25$.}
{In contrast to Atari, Panda therefore shows a smaller numerical effect of the threshold choice, but a clearer distinction between useful and non-useful routing criteria: latent affinity and action affinity are both effective, whereas replay-memory KL and the current hybrid rule are not.}

\begin{table}[H]
\centering
\setlength{\tabcolsep}{3pt}
\resizebox{\textwidth}{!}{%
\begin{tabular}{lccccccc}
\hline
Configuration & Run & Copies & Reach & Push & Pick\&Place & AvgGap & AvgF \\
\hline
TSN-ReplayKL & r1 & 2 & -7.865 & -1.026 & -2.061 & 3.505 & 3.838 \\
TSN-ReplayKL & r2 & 2 & -14.716 & -1.203 & -1.433 & 5.638 & 7.270 \\
TSN-ReplayKL & r3 & 2 & -10.546 & -1.083 & -1.585 & 4.259 & 5.179 \\
\hline
TSN-Affinity-A ($\tau_A=0.65$) & r1 & 3 & -0.188 & -0.956 & -1.505 & 0.737 & 0.000 \\
TSN-Affinity-A ($\tau_A=0.65$) & r2 & 2 & -1.482 & -0.949 & -2.643 & 1.546 & 0.647 \\
TSN-Affinity-A ($\tau_A=0.45$) & r1 & 3 & -0.189 & -0.972 & -1.422 & 0.715 & 0.001 \\
TSN-Affinity-A ($\tau_A=0.35$) & r1 & 3 & -0.189 & -0.972 & -1.422 & 0.715 & 0.001 \\
\hline
TSN-Affinity-L ($\tau_L=25$) & r1 & 3 & -0.189 & -0.949 & -1.308 & 0.670 & 0.000 \\
TSN-Affinity-L ($\tau_L=25$) & r2 & 3 & -0.188 & -0.956 & -1.505 & 0.737 & 0.000 \\
\hline
TSN-Affinity-H ($\tau_H=0.50,\alpha=0.70$) & r1 & 2 & -1.482 & -0.949 & -2.643 & 1.546 & 0.647 \\
\hline
\end{tabular}%
}
\caption{Auxiliary Panda reuse runs. Replay-memory KL routing is weaker and less stable than affinity-based routing.}
\label{tab:app_panda_aux_runs}
\end{table}

\section{Additional Learning Curves}
\label{app:learning_curves}

Each plot shows performance after the end of task $t$.
Background colors indicate the currently trained task, and lines indicate evaluation tasks.
Atari uses single-task-normalized scores.
Panda uses the same normalization only for qualitative comparison.

\subsection{Atari Learning Curves}

Naive is unstable, Cumulative reduces but does not remove backward interference, and the affinity-based TSN variants remain nearly flat after a task has been acquired.

\begin{figure}[!htbp]
    \centering
    \begin{minipage}{0.48\textwidth}
        \centering
        \includegraphics[width=\linewidth]{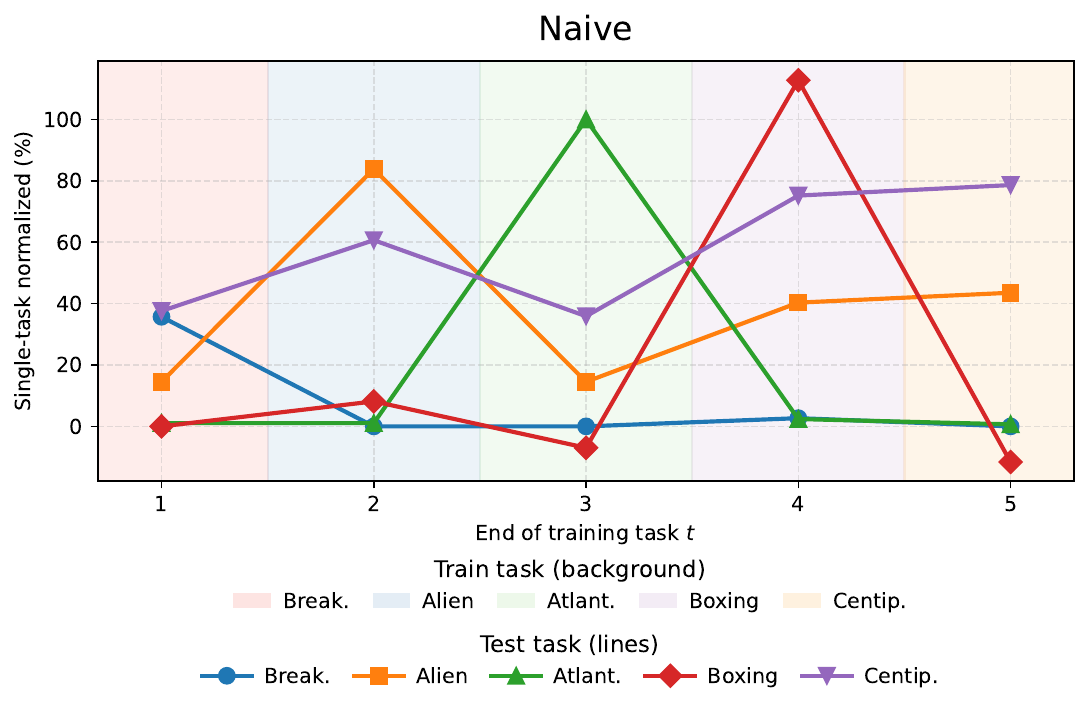}
        \par\small (a) Atari Naive
    \end{minipage}\hfill
    \begin{minipage}{0.48\textwidth}
        \centering
        \includegraphics[width=\linewidth]{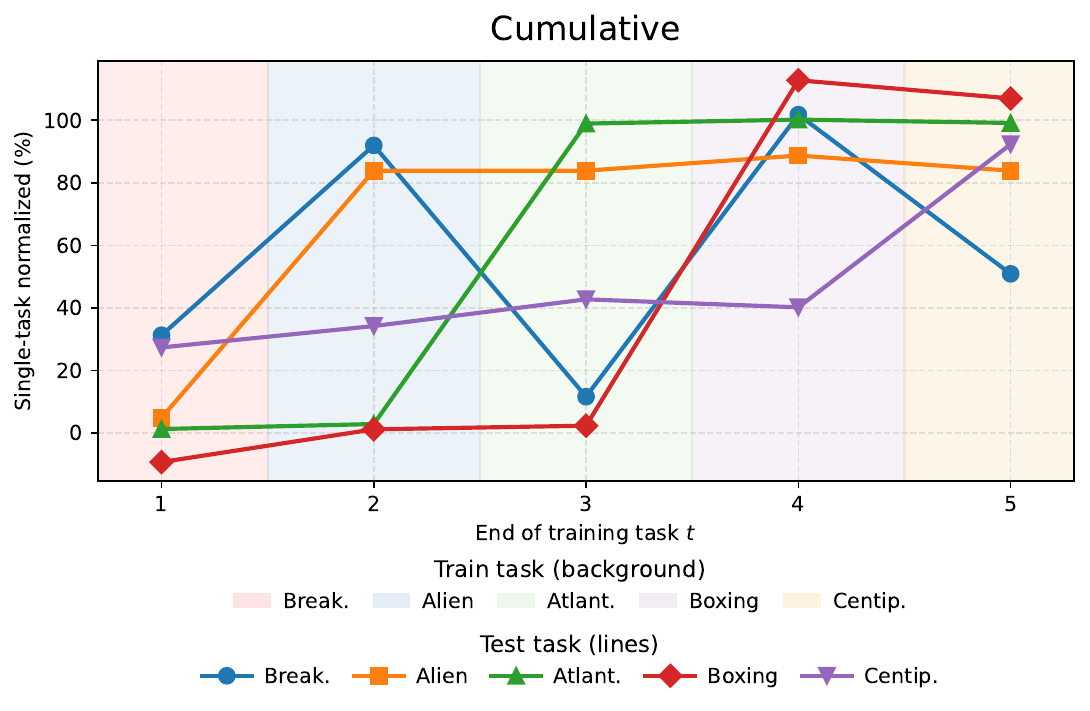}
        \par\small (b) Atari Cumulative
    \end{minipage}

    \vspace{0.5em}

    \begin{minipage}{0.48\textwidth}
        \centering
        \includegraphics[width=\linewidth]{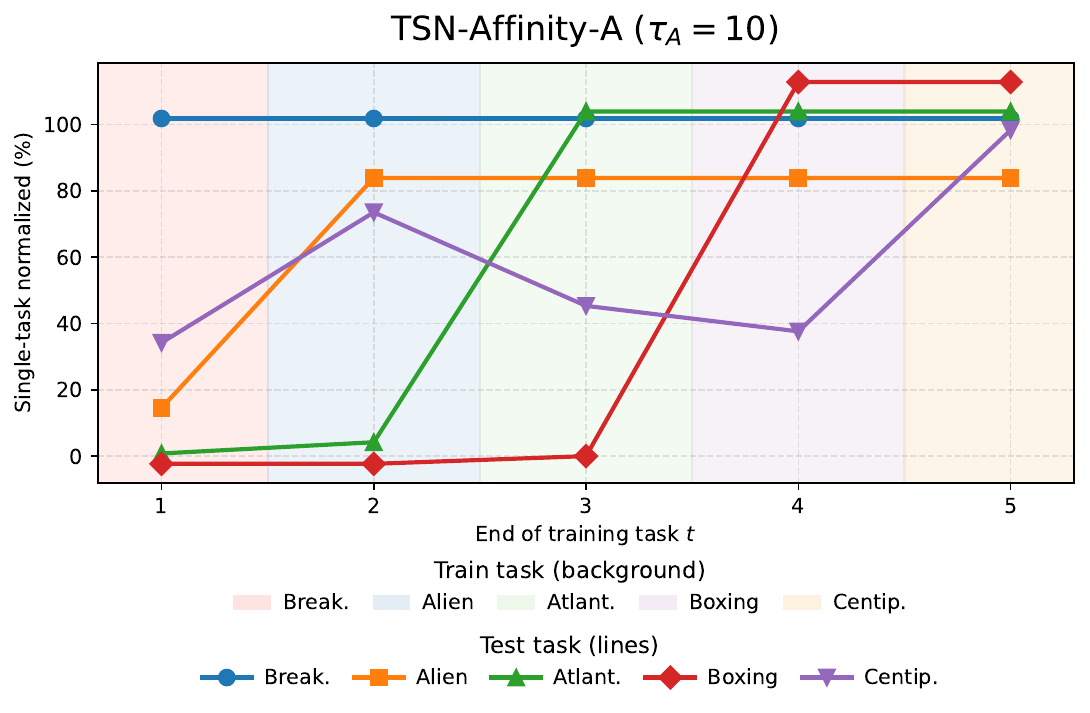}
        \par\small (c) Atari TSN-Affinity-A ($\tau_A=10$)
    \end{minipage}\hfill
    \begin{minipage}{0.48\textwidth}
        \centering
        \includegraphics[width=\linewidth]{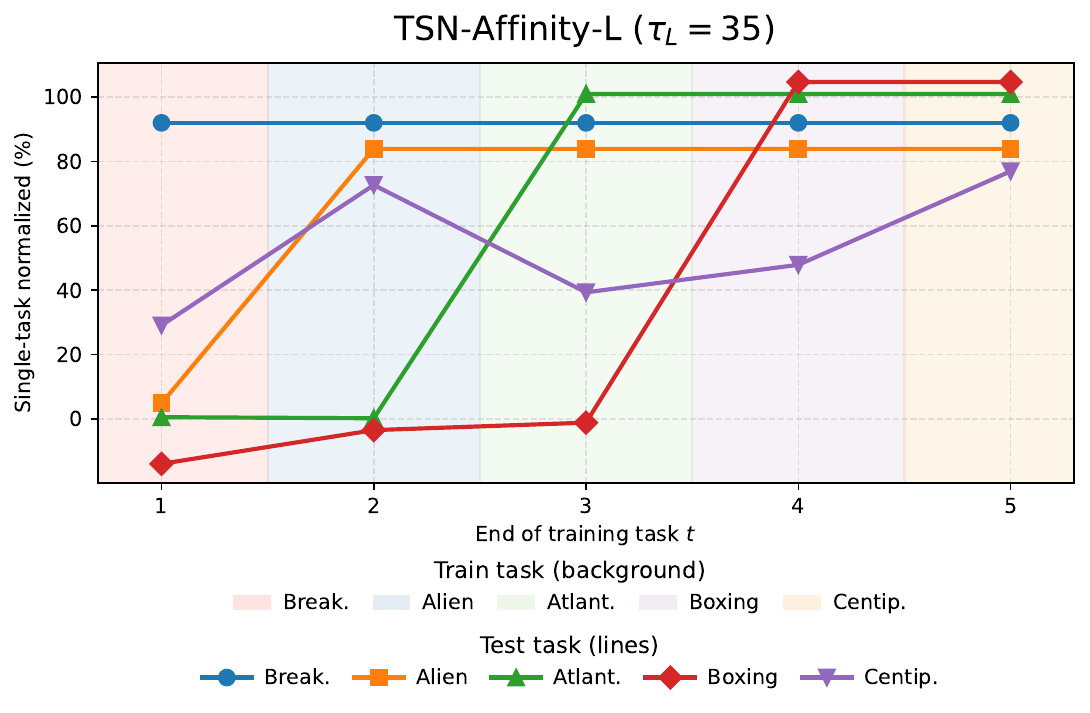}
        \par\small (d) Atari TSN-Affinity-L ($\tau_L=35$)
    \end{minipage}

    \caption{Additional Atari learning curves. The affinity-based TSN variants exhibit the characteristic near-zero-forgetting regime.}
    \label{fig:appendix_atari_curves}
\end{figure}

\subsection{Panda Learning Curves}

Panda is less regular than Atari: Naive remains unstable, Cumulative is more structured, and the affinity-based variants are smoother but not uniformly flat.

\begin{figure}[!htbp]
    \centering
    \begin{minipage}{0.48\textwidth}
        \centering
        \includegraphics[width=\linewidth]{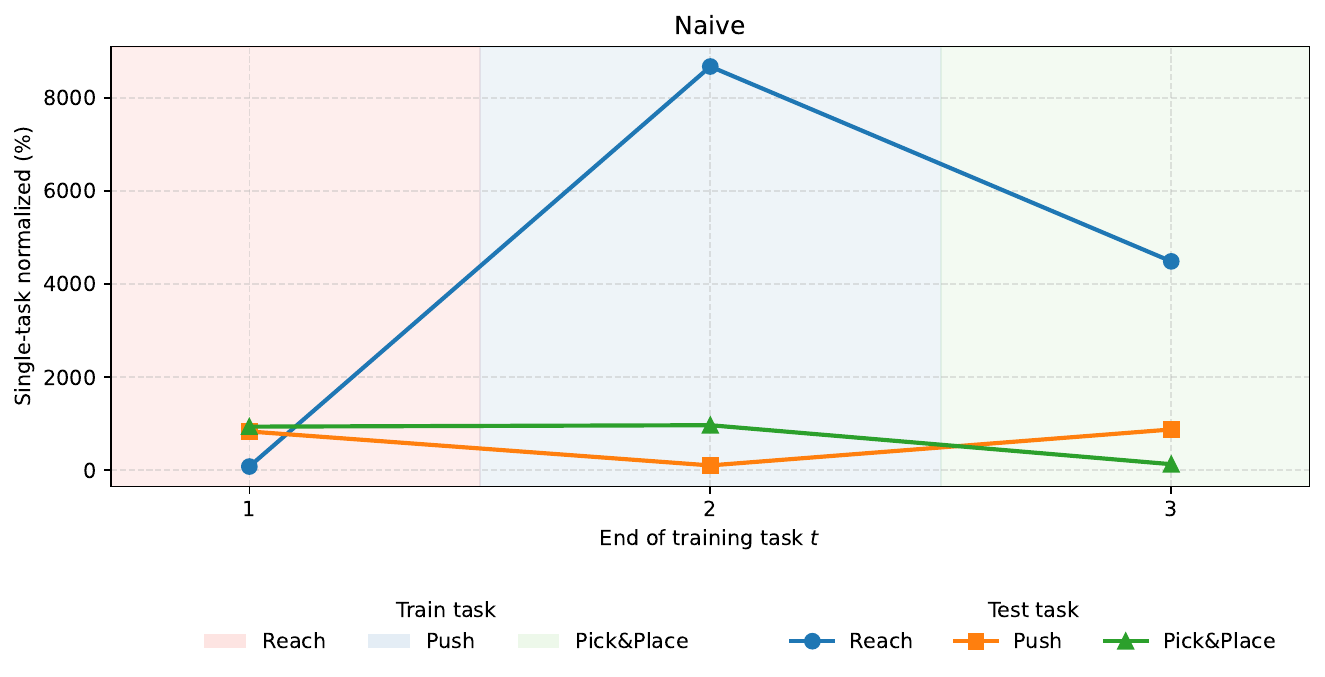}
        \par\small (a) Panda Naive
    \end{minipage}\hfill
    \begin{minipage}{0.48\textwidth}
        \centering
        \includegraphics[width=\linewidth]{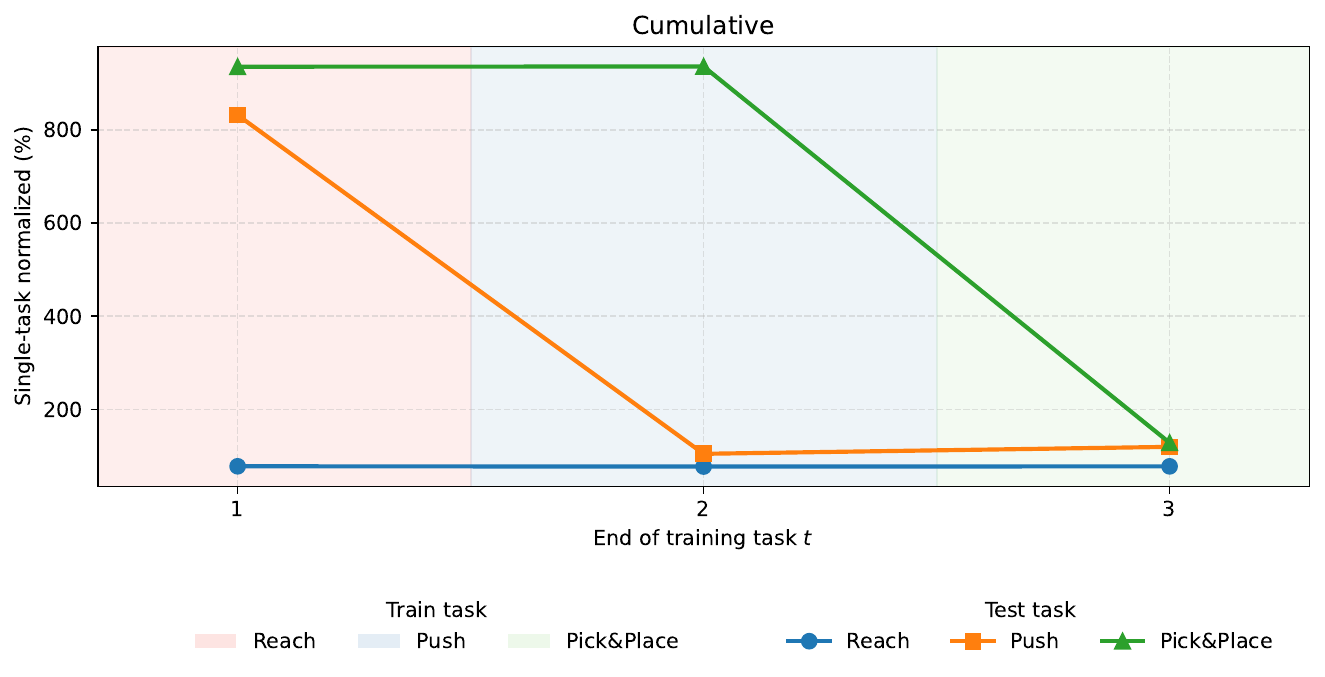}
        \par\small (b) Panda Cumulative
    \end{minipage}

    \vspace{0.5em}

    \begin{minipage}{0.48\textwidth}
        \centering
        \includegraphics[width=\linewidth]{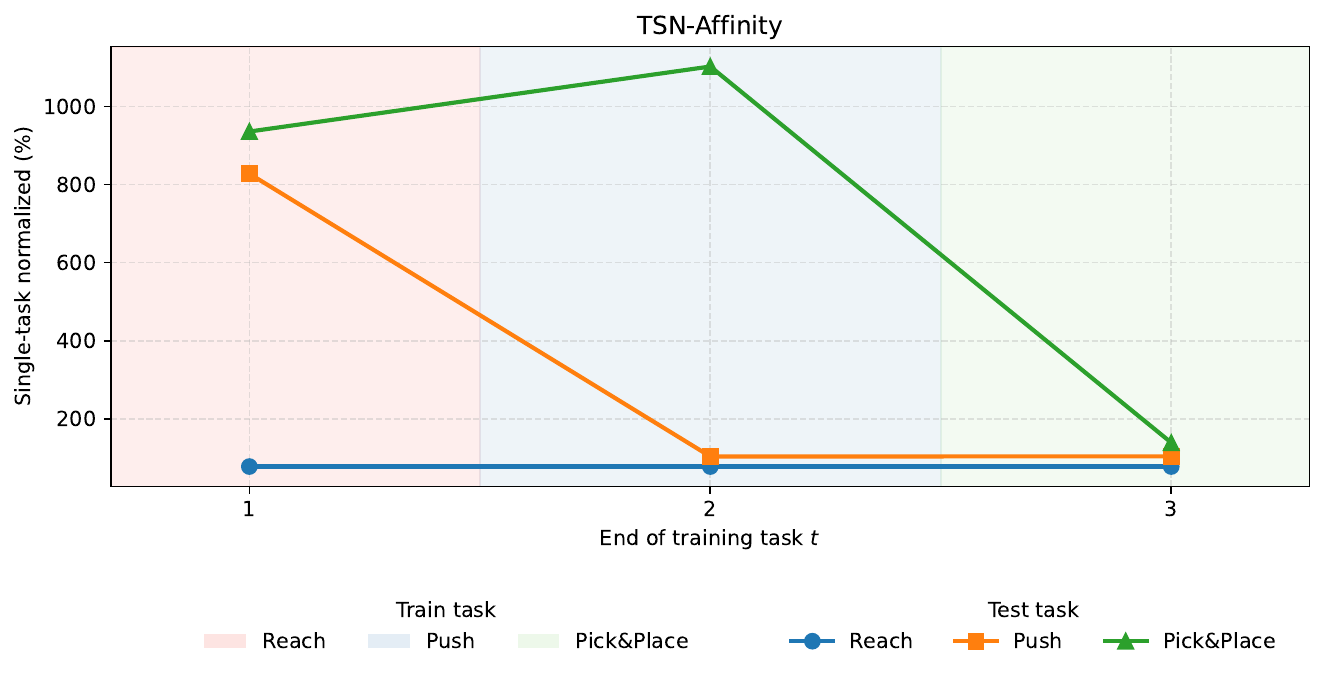}
        \par\small (c) Panda TSN-Affinity-A ($\tau_A=0.45$)
    \end{minipage}\hfill
    \begin{minipage}{0.48\textwidth}
        \centering
        \includegraphics[width=\linewidth]{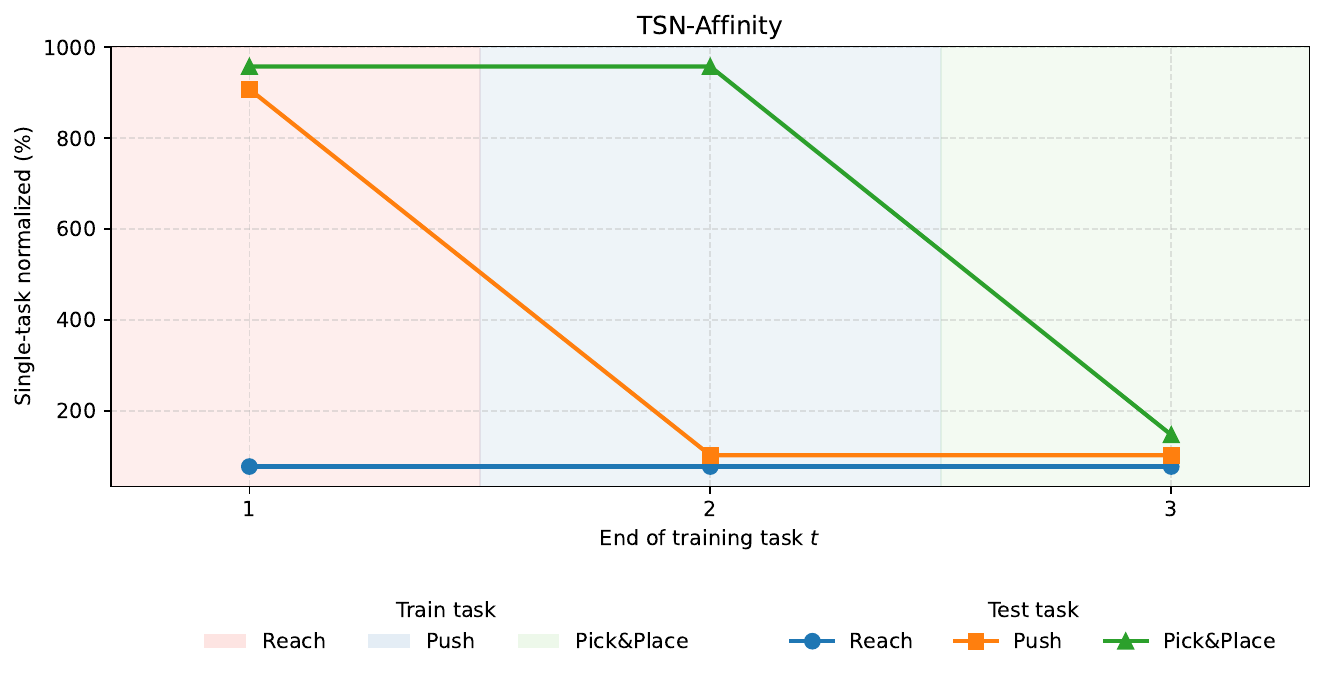}
        \par\small (d) Panda TSN-Affinity-L ($\tau_L=25$)
    \end{minipage}

    \caption{Additional Panda learning curves. These plots are intended mainly for qualitative comparison; the main quantitative discussion in the paper relies on raw returns and average gap-to-target metrics.}
    \label{fig:appendix_panda_curves}
\end{figure}

\section{Panda Padding and Shared Task Interface}
\label{app:panda_padding}
{
The Panda tasks considered in this work differ in their raw observations and action dimensionalities. To support a single continual-learning backbone across the whole task sequence, we embed all tasks into a shared observation space of dimension $26$ and a shared action space of dimension $4$.
For observations, task-specific state vectors are first formed by concatenating \texttt{observation}, \texttt{achieved\_goal}, and \texttt{desired\_goal}, along with an additional normalized time feature used in our environment wrapper. If the resulting vector is shorter than the global observation dimension, it is zero-padded in the trailing coordinates.
For actions, task-specific action vectors are analogously embedded into the shared global action space by zero-padding trailing coordinates. During training, padded action dimensions are excluded from the loss by means of a task-specific binary action mask. Concretely, if task $t$ has an active action mask $m_a^{(t)} \in \{0,1\}^{4}$, then only dimensions with $m_a^{(t)}=1$ contribute to the Panda regression objective. During evaluation, the predicted action vector is truncated back to the original task-specific action dimensionality before being applied in the environment.
This design allows all Panda tasks to share a single Decision Transformer backbone while avoiding artificial penalties on padded coordinates.
}


\end{document}